\documentclass[conference]{IEEEtran}
\IEEEoverridecommandlockouts

\usepackage{cite}

\usepackage[numbers,sort&compress]{natbib}

\usepackage[utf8]{inputenc}
\usepackage{amsmath,amssymb}
\usepackage{graphicx}
\usepackage{bm}
\usepackage{enumerate}
\usepackage{comment}
\usepackage{xcolor}
\usepackage{url}
\usepackage{color,soul}
\usepackage{subcaption}
\usepackage{float}

\definecolor{light-gray}{gray}{0.8}

\usepackage{amsmath,amssymb,amsfonts}
\usepackage{graphicx}
\usepackage{textcomp}
\usepackage{xcolor}
\usepackage{subcaption}

\usepackage{arydshln} 
\usepackage{hyperref}
\usepackage{subcaption} 
\usepackage{caption}
\graphicspath{ {./figs/} }

\def\BibTeX{{\rm B\kern-.05em{\sc i\kern-.025em b}\kern-.08em
    T\kern-.1667em\lower.7ex\hbox{E}\kern-.125emX}}

\begin{document}

\title{Confidence Trigger Detection: Accelerating Real-time Tracking-by-detection Systems

\thanks{Corresponding author: Zhicheng Ding (zhicheng.ding@columbia.edu)}}

\author{
\begin{tabular}[t]{c@{\extracolsep{4em}}c@{\extracolsep{3em}}c} 

1\textsuperscript{st} Zhicheng Ding & 2\textsuperscript{nd} Zhixin Lai & 3\textsuperscript{rd} Siyang Li \\
\textit{Columbia University} & \textit{Cornell University} & \textit{Pace University} \\
New York, USA & Ithaca, USA & New York, USA \\
zhicheng.ding@columbia.edu & zl768@cornell.edu & lisiyang98@hotmail.com \\

\\

4\textsuperscript{th} Panfeng Li & 5\textsuperscript{th} Qikai Yang & 6\textsuperscript{th} Edward Wong \\
\textit{University of Michigan} & \textit{University of Illinois Urbana-Champaign} & \textit{New York University} \\
Ann Arbor, USA & Urbana, USA & Brooklyn, USA \\
pfli@umich.edu & qikaiy2@illinois.edu & ewong@nyu.edu \\
\end{tabular}

}

\maketitle

\begin{abstract}
Real-time object tracking necessitates a delicate balance between speed and accuracy, a challenge exacerbated by the computational demands of deep learning methods. In this paper, we introduce Confidence-Triggered Detection (CTD), a novel approach that strategically skips object detection for frames exhibiting high similarity, leveraging tracker confidence scores. CTD not only enhances tracking speed but also preserves accuracy, surpassing existing tracking algorithms. Through extensive evaluation across various tracker confidence thresholds, we identify an optimal trade-off between tracking speed and accuracy, providing crucial insights for parameter fine-tuning and enhancing CTD's practicality in real-world scenarios. Furthermore, our experiments across diverse detection models underscore the robustness and versatility of the CTD framework, demonstrating its potential to enable real-time tracking in resource-constrained environments.

\end{abstract}


\section{Introduction}
\label{sec:intro}

The field of object detection has witnessed a dramatic shift in recent years, with deep learning architectures achieving unprecedented performance. However, video analysis presents a unique challenge that goes beyond the analysis of isolated frames. Unlike static images, videos possess an inherent temporal dimension, where the relationships between frames hold crucial information. Capturing these temporal dynamics is essential for a comprehensive understanding of the visual content in videos \cite{Bewley2016,Leal-Taixe2016,Wojke2018,Xiang2015}. This is particularly important in the context of multiple object tracking (MOT), which has numerous applications in surveillance systems, intelligent robotics, and autonomous vehicles \cite{Huang2015,Zaman_2023}.

Most recent MOT approaches leverage a tracking-by-detection paradigm \cite{Bouchrika2016}. This involves three key steps: (1) object detection in each frame, (2) object tracking based on information from previous frames, and (3) data association using location and feature data from both detector and tracker \cite{Wojke2017simple}. While algorithms like SORT perform well in terms of precision and accuracy, they often suffer from ID switches and struggle with occlusions. DeepSORT addresses this with a better association metric, but real-world deployment necessitates both high accuracy and real-time performance \cite{Wojke2018}. 

Despite the surge in proposed MOT algorithms, achieving real-time performance remains a significant challenge~\cite{peng2024maxk}. The MOTChallenge leaderboard exemplifies this inherent trade-off: faster (higher Hz) algorithms typically exhibit lower Multiple Object Tracking Accuracy (MOTA), while higher MOTA results tend to be associated with slower processing speeds.

Frame skipping is a commonly used technique to improve speed. It involves processing only a subset of video frames, either by skipping a fixed number or waiting for the tracker to be ready for the next assignment. While this reduces computational load and improves speed, it comes at the cost of reduced tracking accuracy. Critical information might be missed, leading to tracking errors such as objects shifting significantly or being entirely lost.

Inspired by frame skipping, this paper presents a novel Confidence-Triggered Detection (CTD) approach. CTD leverages the confidence score associated with object association to determine when to trigger object detection. This allows the system to strategically skip frames where objects exhibit minimal movement, minimizing unnecessary computations. Conversely, a significant discrepancy between the tracker's predicted location and the detector's output triggers a new detection. This approach aims to achieve a balance between processing speed and tracking accuracy compared to traditional fixed-interval frame skipping.

In summary, our paper makes the following primary contributions:
\begin{itemize}
    \item [$\bullet$] We present CTD, a novel general real-time tracking framework designed to enhance tracking speed while maintaining high accuracy.

    \item [$\bullet$] Our evaluation of the CTD framework under varying confidence thresholds reveals an optimal tradeoff between tracking speed and accuracy. This assessment provides critical insights for tuning CTD parameters to suit specific tracking contexts, thereby bolstering its practicality and performance in real-world deployments.

    \item [$\bullet$] We perform extensive experiments to assess the performance of CTD across a range of detection models which demonstrate the robustness and versatility of the CTD framework. 

\end{itemize}

\section{Related Work}
This section discusses relevant research areas related to multiple object tracking (MOT), including tracking-by-detection paradigms, real-time MOT performance considerations, lightweight object detectors, and the Deep SORT algorithm.

\subsection{Tracking-by-detection}
The dominant paradigm in MOT is tracking-by-detection \cite{Breitenstein2011}. This approach leverages two key components: an object detector and a tracker. In each frame, the detector identifies and localizes objects. The tracker then predicts the current location of previously detected objects based on their past states. Finally, an association step links detected objects from the current frame with corresponding tracks using information from both the detector and tracker \cite{Wojke2018}. This approach offers improved accuracy compared to alternatives, as the tracker's predictions can be continuously refined by up-to-date detection.

\subsection{Real-time Object Tracking}

Defining a universal speed threshold for real-time performance in MOT is challenging due to variations in video frame rates captured by different cameras. However, a system is generally considered real-time if its processing speed surpasses the input video frame rate \cite{Kuo2013}. If the processing speed falls behind, a delay accumulates, causing the system output to become out of sync with the actual video and events.

Frame skipping is a common way for real-time object tracking systems where the system processes only a subset of video frames. One approach involves skipping a fixed number of frames after each tracking assignment. However, this method can lead to missing critical frames, especially for fast-moving objects, resulting in significant accuracy drops.

Our proposed CTD approach also incorporates frame skipping. However, unlike fixed-interval skipping, CTD leverages a confidence score to selectively trigger new detection. This allows the system to strategically skip frames with minimal object movement while maintaining tracking accuracy. We continuously run the tracker in all frames to monitor the discrepancy between the tracker's predictions and the latest detection. This difference serves as a measure of confidence, and a significant discrepancy triggers a new detection to correct potential biases before they accumulate significantly. This approach aims to achieve a balance between processing speed and tracking accuracy compared to traditional fixed-interval frame skipping.

\subsection{Lightweight Detectors}
Deep learning-based lightweight detectors are designed to have a smaller number of parameters or require fewer computational passes compared to their heavier counterparts. While this translates to faster processing speeds, it typically comes at the expense of accuracy. In this paper, we utilize the YOLOv3 Tiny, MobileNet-SSD and SqueezeNet for our experienment. 

\subsubsection{YOLOv3 Tiny}
 YOLOv3 Tiny is a reduced version of the YOLOv3 detector. It utilizes a simpler backbone network for feature extraction but still incorporates a Feature Pyramid Network (FPN) to improve the detection of small objects \cite{Redmon2018}. Since YOLOv3 Tiny performs object detection in a single pass, it boasts a significant speed advantage compared to heavier detectors like R-CNN \cite{Girshick2014}). While YOLOv3 achieves the highest MOTA (34\%) in our experiments, its frame rate is limited to 10.915 Hz.

\subsubsection{MobileNet-SSD}
MobileNets are neural networks that perform efficiently. It can also be used on mobile devices and reach a fairly high accuracy \cite{Howard2017}. SSD \cite{Liu2016} uses VGG16 \cite{Simonyan2015} to extract feature maps. SSD classifies and locates objects in a single forward pass. MobileNet-SSD combines SSD and MobileNets which perform high speed and relatively high accuracy. Using MobileNet SSD, the system reaches 17.5\% MOTA and performs 15.604 Hz of speed.

\subsubsection{SqueezeNet}
SqueezeNet uses a squeeze layer and an expanded layer to reach a really fast performance. The paper on SqueezeNet provides a quantitative analysis to show that SqueezeNet can be 510 times fewer parameters to reach the same accuracy as AlexNet \cite{Krizhevsky2012}. Using SqueezeNet, the system reaches the lowest MOTA (9.4\%) but performs the fastest (21.673 Hz).

\subsection{DeepSORT}
DeepSORT \cite{Wojke2018} is an extension of the SORT algorithm \cite{Bewley2016} that incorporates appearance information for object matching. This enables Deep SORT to track objects even during extended occlusions. Additionally, it utilizes the Mahalanobis distance metric \cite{Chang2014,smalheiser2021effect} to incorporate motion information into the association process.

We leverage DeepSORT for two main reasons in our approach. Firstly, DeepSORT can track objects during occlusions ensuring that objects can still be matched after skipping a certain number of frames. Secondly, we utilize the Mahalanobis distance calculated by Deep SORT to infer low confidence scores and trigger new detections when necessary.


\section{Method}
\label{sec:method}

This section details the methodology employed in our object-tracking system. We first provide a high-level overview of the tracking-by-detection framework utilized. Then, we explain the Confidence-Triggered Detection (CTD) approach, a core component responsible for strategically skipping object detection for enhanced processing speed in greater detail. Finally, we explain how the system infers the confidence score using Mahalanobis distance.

\subsection{Tracking-by-detection Framework} \label{Tracking-by-detection Framework}

Our system leverages a tracking-by-detection framework, as illustrated in Figure~\ref{fig:frames}. This framework shares similarities with DeepSORT \cite{Wojke2018} and integrates three key components: Kalman Filter, Hungarian Assignment, and the CTD module.

\noindent\textbf{Kalman Filter:} When a new frame arrives, the Kalman Filter predicts the object's location in the current frame based on its detection information from the previous frame. This prediction helps maintain object tracking during periods when detection is skipped.

\noindent\textbf{Hungarian Assignment:}  This algorithm plays a crucial role in object association and ID attribution. It essentially determines whether an object detected in the current frame corresponds to the same object tracked in the previous frame.

\noindent\textbf{Confidence-Triggered Detection (CTD):} This innovative method integrates the Mahalanobis Distance and a derived Confidence Score with Detector Trigger Modules for object detection. The Confidence Score, deduced from the Mahalanobis Distance using a chi-squared distribution, dictates whether to maintain tracking or invoke the detector for new object detection, as elaborated in Section \ref{Confidence Inference}. Specifically, when the confidence score falls below a predefined threshold or the number of consecutively skipped frames exceeds a predetermined threshold, new object detection will be triggered and the previous detection result will be replaced. Otherwise, object detection is skipped for this frame. The intricacies of the CTD approach, including detector operations and decision mechanisms, are further expounded in Section \ref{Confidence Trigger Detection (CTD) Approach}.


\subsection{Confidence-Triggered Detection (CTD) Approach} \label{Confidence Trigger Detection (CTD) Approach}

The CTD approach aims to achieve a balance between processing speed and tracking accuracy by strategically skipping object detection in certain frames. Here's how it operates:

\noindent\textbf{Initialization:} The system starts with a counter set to a value exceeding the maximum frame-skipping threshold. Additionally, the confidence score is initialized to 0\%. The initial settings ensure object detection occurs in the first frame and establish initial object locations.

\noindent\textbf{Confidence Score Evaluation:} The Mahalanobis Distance is calculated between two bounding boxes. The predicted bounding of the current frame by the Kalman filter and bounding boxes of the last frame are detected by the detector. By applying a chi-squared distribution to the Mahalanobis Distance, as outlined in Section~\ref{Confidence Inference}, we obtain a confidence score, which will be used to determine if triggering a new round of detection in the Detection Trigger Module.

\noindent\textbf{Detection Trigger Module:} If the confidence score surpasses a predefined threshold and the frame-skipping counter hasn't reached its limit, the system skips detection and relies solely on the Kalman filter's prediction for the current frame (detection is skipped). This approach improves processing speed. Conversely, if the confidence score falls below the threshold or the frame-skipping counter reaches its limit, a new detection is triggered in the current frame to potentially correct the discrepancies between the predicted and actual object location.

\noindent\textbf{Data Association:} The Hungarian algorithm is employed for the data association task, effectively matching the detection results from the current frame with the predicted object locations provided by the Kalman filter. This process ascertains the continuity of object identity between consecutive frames, ensuring that each detected object is accurately aligned with its corresponding track.

By strategically skipping detection based on confidence scores, the CTD approach achieves a desirable balance between processing speed and tracking accuracy.

\subsection{Confidence Score Inference} \label{Confidence Inference}

The Mahalanobis Distance is a metric used to assess the similarity between a data point and a distribution defined by its mean and covariance. In the context of multi-object tracking, it can be employed to measure the discrepancy between a predicted bounding box and the detected objects in a video frame, defined as:

\begin{equation}
M(i, j)^2 = (x_j - \hat{y_i})^T (S_i)^{-1} (x_j - \hat{y_i})
\end{equation}

\noindent where $i$ denotes the information at the $i$-th frame and $j$ denotes the information at $j$-th frame. The $x_j$ is the matrix of all the predicted bounding boxes at the $j$-th frame. The $y_i$ and $S_i$ represent the mean and covariance of detected bounding boxes. The Mahalanobis distance follows a chi-square distribution \cite{Chang2014}. Given probability level $p$ and degrees of freedom $v$, we can calculate the distance threshold corresponding to that probability using the inverse cumulative distribution function (CDF) \cite{Pearson1992,Wojke2018,wang2023embracing,tang2024zerothorder,wang2021deminet,ding-24-style}:

\begin{equation}
d = F^{-1}(p|v)
\end{equation}

\noindent where $p$ can be calculated using CDF, shown as equation (3).

\begin{equation}
p = F(x|v) = \int_0^x \frac{t^{(v-2)/2}\exp({x/2})}{2^{v/2}\Gamma(v/2)}dt
\end{equation}

\begin{figure*} [ht]
  \centering
  \begin{minipage}{.14\textwidth}
    \centering
    \includegraphics[width=\linewidth]{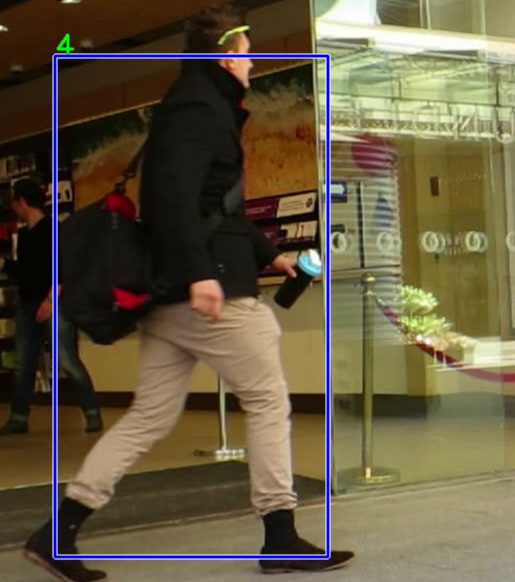} \\ \footnotesize{(a) Frame 55}
  \end{minipage}\hfill%
  \begin{minipage}{.14\textwidth}
    \centering
    \includegraphics[width=\linewidth]{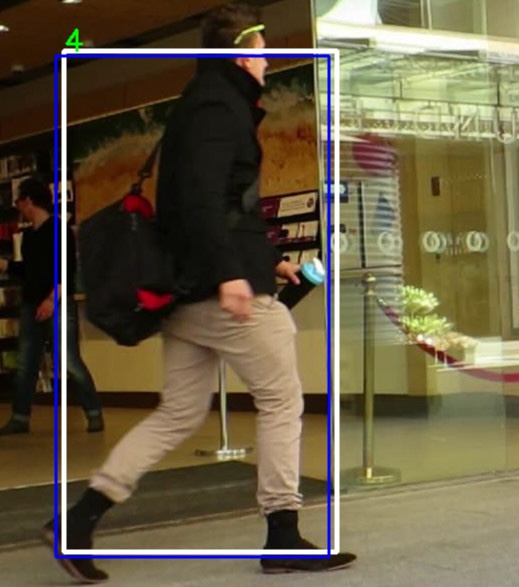} \\ \footnotesize{(b) Frame 56}
  \end{minipage}\hfill%
  \begin{minipage}{.14\textwidth}
    \centering
    \includegraphics[width=\linewidth]{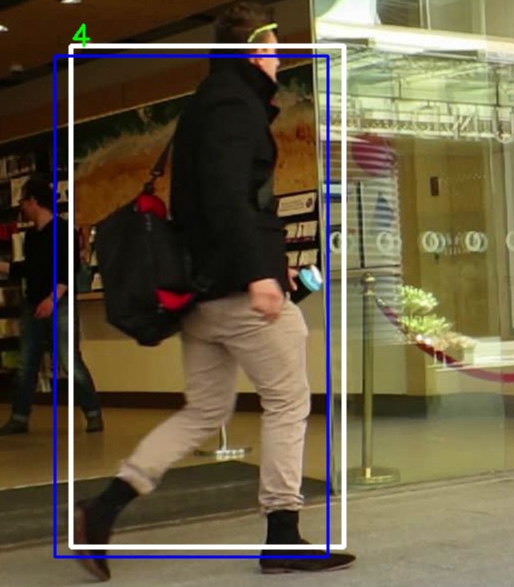} \\ \footnotesize{(c) Frame 57}
  \end{minipage}\hfill%
  \begin{minipage}{.14\textwidth}
    \centering
    \includegraphics[width=\linewidth]{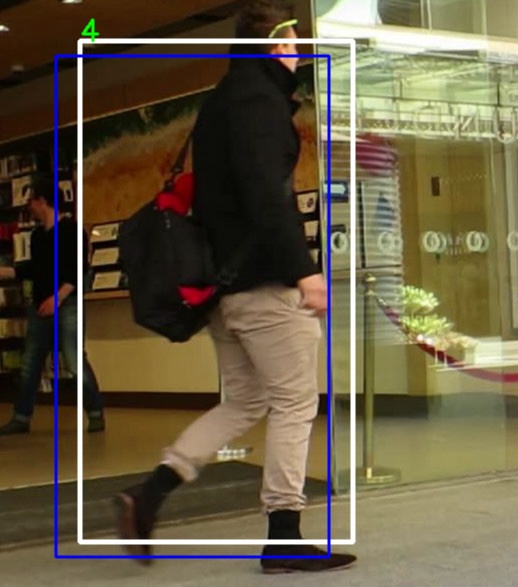} \\ \footnotesize{(d) Frame 58}
  \end{minipage}\hfill%
  \begin{minipage}{.14\textwidth}
    \centering
    \includegraphics[width=\linewidth]{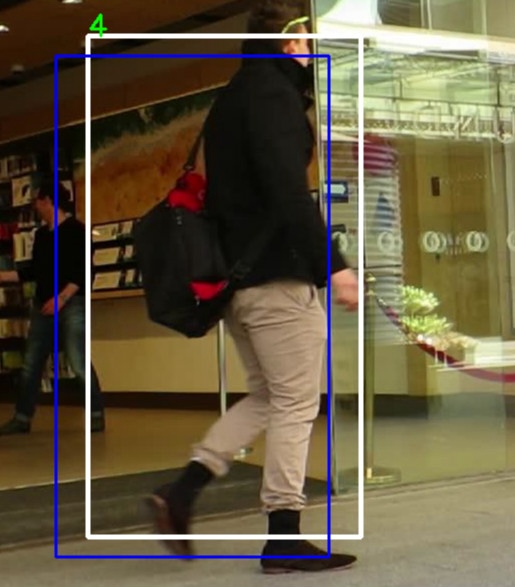} \\ \footnotesize{(e) Frame 59}
  \end{minipage}\hfill%
  \begin{minipage}{.14\textwidth}
    \centering
    \includegraphics[width=\linewidth]{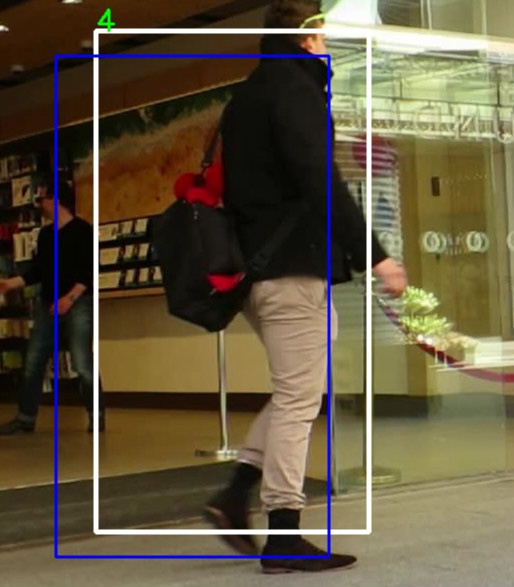} \\ \footnotesize{(f) Frame 60}
  \end{minipage}\hfill%
  \begin{minipage}{.14\textwidth}
    \centering
    \includegraphics[width=\linewidth]{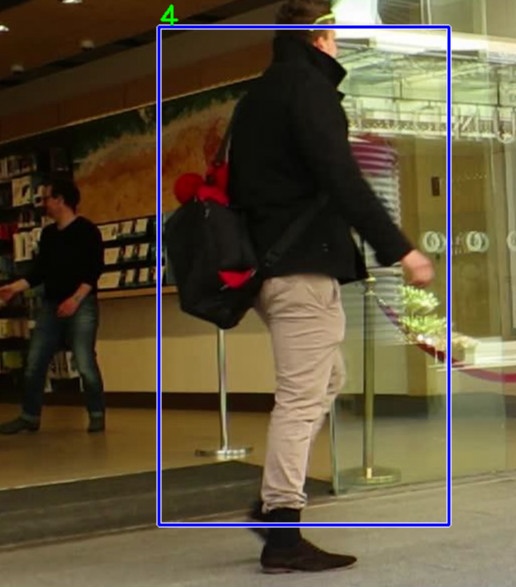} \\ \footnotesize{(g) Frame 61}
  \end{minipage}%
  \caption{illustrates the tracking outcomes for frames 55 to 61 utilizing the CTD approach. The analysis is illustrated with cropped image frames, where white bounding boxes denote predictions made by the Kalman filter, and blue bounding boxes indicate detections from the object detector. Between frames 56 and 60, no new detections were recorded, because the confidence score, derived from the comparison of the white and blue bounding boxes, exceeded the predetermined threshold. However, at frame 61, a new detection was prompted due to the confidence score falling below the threshold. This indicates a notable deviation between the prediction from the Kalman filter and the object detection results from the previous frame.}
  \label{fig:frames}
  \end{figure*}

\noindent where $\Gamma(\cdot)$ represents the Gamma function. In our paper, we set the degrees of freedom $v$ to $4$. We evaluate the model's performance using different probability values $p$.

\section{Experiment}
\label{sec:experiment}

\begin{table}[htbp]
\begin{center}
\setlength{\tabcolsep}{4pt}
\caption{Comparison of 2D MOT results of CTD and existing models. The method above the dashed line is an on-skippable approach that makes detection to each frame, while the methods below the dashed line are skippable methods that selectively skip frames for detection. The table presents the performance metrics, with the best values of skippable methods highlighted in \textbf{bold}. CTD achieves the best accuracy (MOTA) and Speed among all the skippable methods. }
\begin{tabular}{ l | c   c   c   c   c   c } 
\textbf{Method}  & \textbf{MOTA $\uparrow$} & \textbf{FP $\downarrow$} & \textbf{Speed (Hz) $\uparrow$} \\
\hline
TraByDetNs                  & 22.2 & 4,664 & 8.1
\\
\hdashline
TC\_SIAMESE      & 19.3 & 6,127 & 13.0
\\
TSDA\_OAL                    &18.6 & 16,350 & 19.7
\\
GMPHD                       &18.5 & 7,864 & 19.8
\\
LDCT                        &4.7 & 14,066 & 20.7
\\
Ours(CTD)                   & \textbf{19.5} & \textbf{5,115} & \textbf{21.4}
\\
\end{tabular}
\label{table:sota1}
\end{center}
\end{table}

\begin{table*}[]
\centering
\small
\setlength{\tabcolsep}{4pt}
\caption{In the context of the CTD framework, we conducted a performance comparison across various detectors, setting the maximum frame detection limit to 8 and employing a baseline confidence threshold of 100\%. To guarantee uniform testing conditions, all detectors were evaluated on identical 2D video clips sourced from the MOTChallenge dataset. From the results, the consistent speed-accuracy balance is maintained across all models when incorporated with CTD.}
\begin{tabular}{l | c c c c c c}
                     & \multicolumn{2}{c}{YOLOv3 Tiny} & \multicolumn{2}{c}{MobileNet} & \multicolumn{2}{c}{SqueezeNet} \\
\hline
Confidence Threshold & Speed Gain    & Accuracy Lose   & Speed Gain   & Accuracy Lose  & Speed Gain    & Accuracy Lose  \\
\hdashline
100\% (baseline)     & 0             & 0               & 0            & 0              & 0             & 0                \\
90\%                 & 29.2\%    & -2.1\%    & 29.4\%  & -3.0\%          & 26.6\%   & -2.8\%   \\
10\%                 & 148.1\%   & -26.2\%    & 145.5\%  & -28.1\%          & 141.3\%   & -27.8\%  
\end{tabular}
\label{table:diff-detector}
\end{table*}

This section assesses the efficacy of the CTD approach in improving real-time object-tracking performance. We detail our experimental setup, encompassing datasets, models, and frame-by-frame analysis. Then we conduct comparative assessments against existing techniques and demonstrate the superior performance of the CTD. Furthermore, we analyze the accuracy-speed trade-off across different confidence score thresholds, which yields valuable insights into optimizing CTD's parameters for different tracking scenarios. Ultimately, we perform extensive experiments to assess the performance of CTD across a range of detection models which demonstrates the robustness and versatility of the CTD framework.

\subsection{System Setup}
We developed a tracking-by-detection system following the framework outlined in Section \ref{Tracking-by-detection Framework}. DeepSORT was chosen as the tracker due to its real-time capabilities, occlusion handling, and competitive MOTA scores \cite{Wojke2018}. YOLOv3 Tiny \cite{Redmon2018} served as the object detector due to its balance between speed and accuracy. DeepSORT integrated with YOLOv3 Tiny achieved the best processing speeds compared to the existing algorithms.

We employed 2D video clips sourced from the MOTChallenge dataset \cite{MOTChallenge2015}. This dataset offers a comprehensive range of challenges, featuring diverse target motions, camera perspectives, and pedestrian densities, rendering it ideal for rigorous evaluation. 

\subsection{CTD Tracking: Frame-by-Frame Analysis}
An in-depth analysis was conducted on frames 55 to 61 from a video within our training set, as depicted in Figure~\ref{fig:frames}. For enhanced visual clarity, we have utilized white bounding boxes to denote the predictions from the Kalman filter and blue bounding boxes to indicate the detector's output. Additionally, the frames have been cropped to concentrate on the analysis of a singular object of interest.

\begin{enumerate}
\item \textbf{Trigger Detection (Frame 55)}: A new detection is triggered at frame 55 due to the low confidence score. A new white bounding box is then generated by the Kalman filter based on the new detection result. Following this detection, the confidence score notably improves, reaching a high level.

\item \textbf{High Confidence Frames (Frames 56-60)}: Despite the man's movement causing slight shifts in the white bounding box, the confidence score consistently exceeds the predetermined threshold. Consequently, detection is omitted, and the blue bounding box remains stationary. However, it's noteworthy that during this period, the confidence score experienced a decline due to the increasing misalignment between the white and blue bounding boxes.

\item \textbf{Trigger Detection (Frame 61)}: The confidence score falls below the predefined threshold, and thus a new detection is activated, leading to an update of the detector's blue bounding box. Consequently, the white bounding box generated by the Kalman filter dynamically adjusts to align with the updated blue bounding box.

\end{enumerate}

Illustrated by the bounding box movements spanning frames 55 to 61, the CTD effectively enhances system speed by selectively skipping video frames during periods of high confidence. 

\subsection{Comparison with Existing Methods}

Following an exhaustive exploration of multiple confidence score thresholds, we determined that establishing the threshold at 30\% achieves a commendable equilibrium between precision and computational efficiency. Subsequently, we conducted comparative analyses between our proposed method CTD and existing non-skippable and skippable tracking approaches. The results are summarized in Table \ref{table:sota1}. As a non-skippable approach, TraByDetNs operates object detection for each frame, contrasting with skippable methods such as TSDA\_OAL \cite{TSDA_OAL}, TC\_SIAMESE \cite{TC_SIAMESE}, and GMPHD \cite{GMPHD}, which perform detection intermittently across frames. Our proposed method, CTD, outperforms all evaluated methods in terms of processing speed and achieves superior MOTA (Multi-Object Tracking Accuracy), and FP (False Positive) metrics compared to other skippable methods.


\begin{figure*}
  \centering
  \begin{subfigure}[T]{0.3\textwidth}
     \includegraphics[width=\columnwidth]{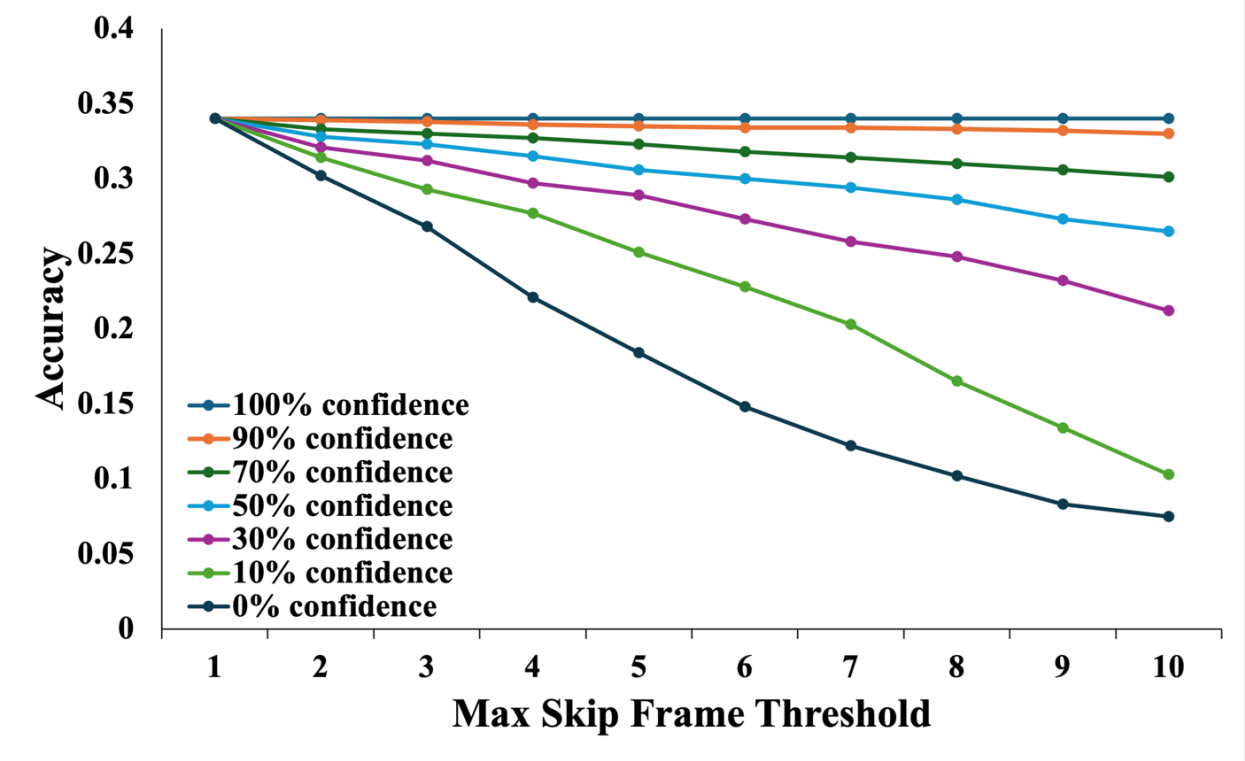}
     \subcaption{\footnotesize{Correlation between Accuracy and Max Skip Frame Threshold}}
  \end{subfigure}
  ~
  \begin{subfigure}[T]{0.3\textwidth}
     \includegraphics[width=\columnwidth]{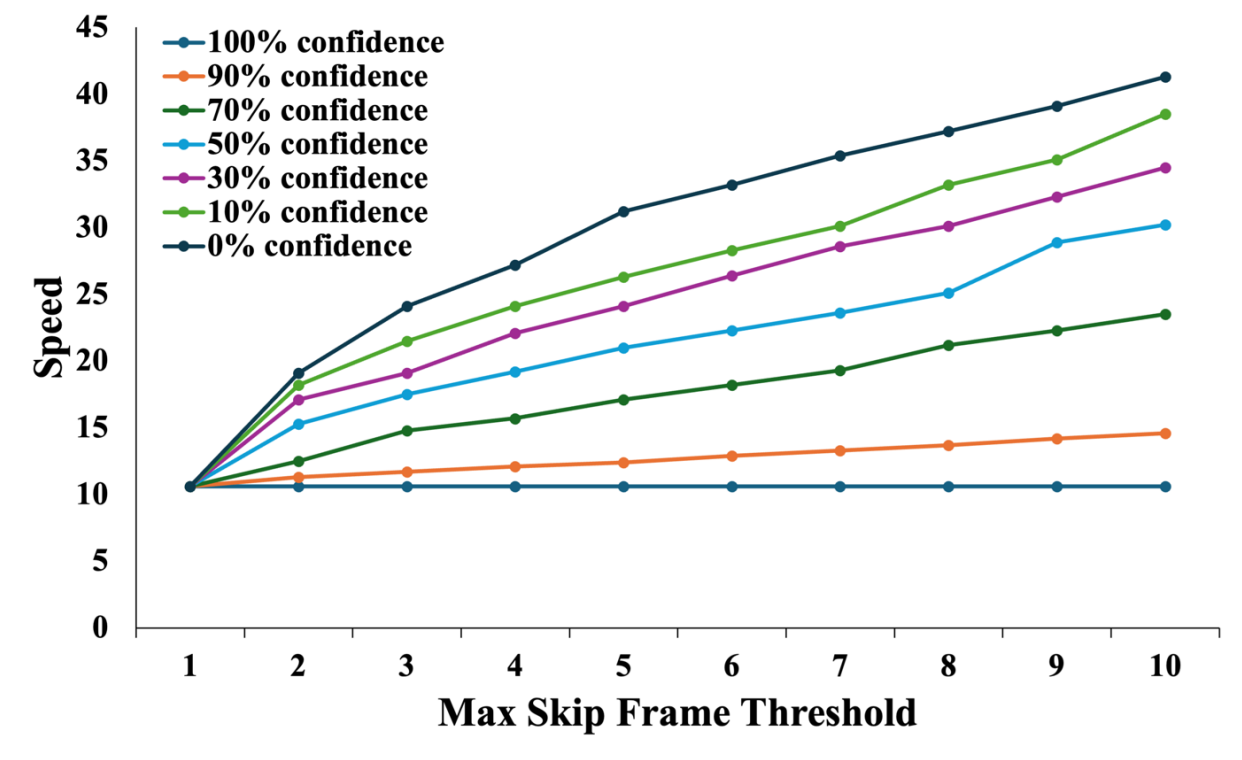}
     \subcaption{\footnotesize{Correlation between Speed and Max Skip Frame Threshold}}
  \end{subfigure}
  ~
  \begin{subfigure}[T]{0.3\textwidth}
     \includegraphics[width=\columnwidth]{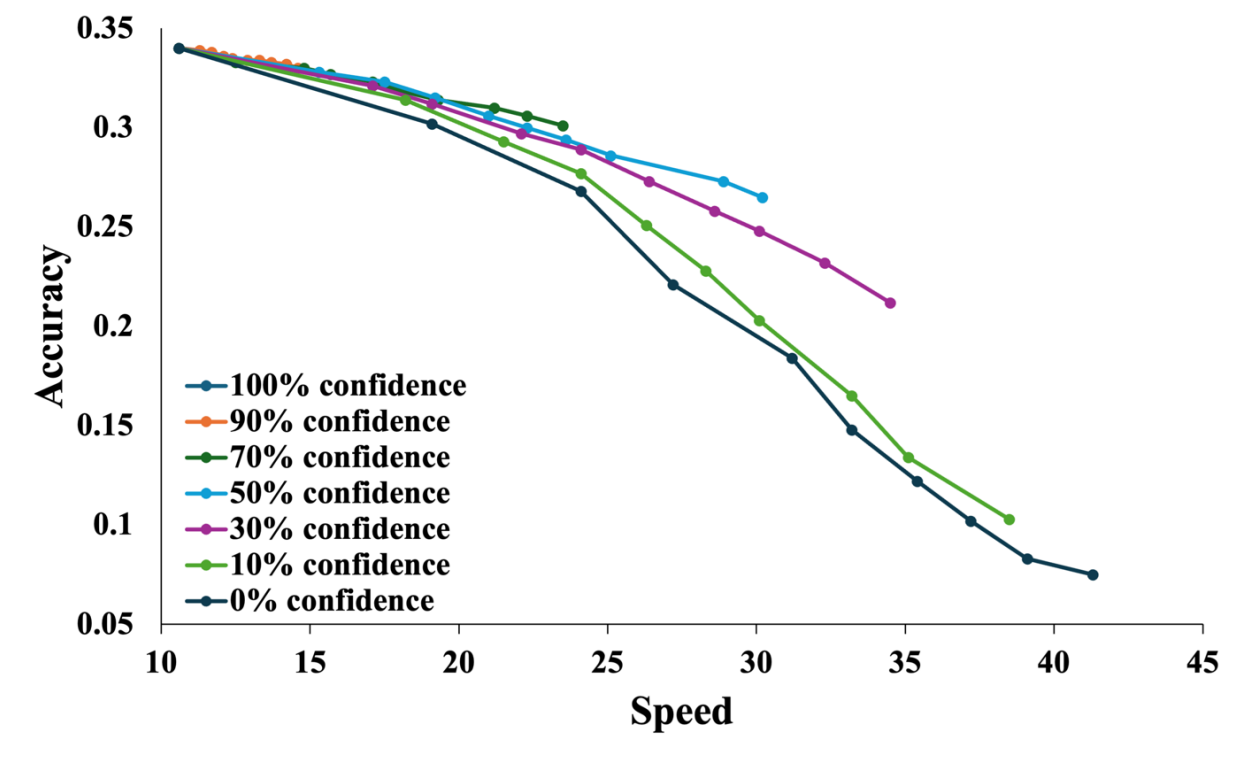}
      \subcaption{\footnotesize{Correlation between Accuracy and Speed}}
  \end{subfigure}

   \label{fig:acc_speed}

   \caption{illustrates the trade-off between speed (FPS) and accuracy (MOTA) using different confidence thresholds.\textbf{100\% Threshold:} Require detection for each frame, considered as baseline. \textbf{0\% Threshold:} Same as fixed frame skipping, never triggers detection based on confidence score but triggers when reaching the maximum skip frame threshold. It highlights that CTD consistently achieves superior accuracy levels at the same speed compared to the fixed frame skipping strategy (0\% confidence threshold).}
\end{figure*}

\subsection{Trade-Off Between Accuracy and Speed}

Although the CTD method enhances overall processing speed by judiciously skipping object detection. However, this optimization strategy may potentially compromise tracking accuracy (MOTA). To thoroughly assess the performance of CTD, we conducted evaluations using various confidence thresholds. This allowed us to delve into the nuanced trade-off between speed (FPS) and accuracy (MOTA) across diverse scenarios. Specifically, we varied the maximum skip frame threshold from 1 to 10 and explored confidence thresholds ranging from 100\% down to 0\%.

Notably, we established two extreme cases for threshold settings to provide additional context for our evaluation. A 100\% threshold mandates detection in each frame, representing the baseline condition. On the other hand, the 0\% threshold mirrors fixed frame skipping, whereby detection is solely triggered based on reaching the maximum skip frame threshold, disregarding confidence scores.

In Figures 3(a) and 3(b), we present the variations in processing speed (FPS) and accuracy (MOTA) under varying confidence thresholds. As the confidence threshold is elevated, the incremental speed advantage is reduced, and correspondingly, the detriment to accuracy becomes less pronounced. This observation suggests a trade-off between speed and accuracy that is modulated by the confidence threshold, with higher thresholds diminishing returns in speed but offering a buffering effect against accuracy degradation.

Figure 3(c) illustrates the direct relationship between speed (FPS) and accuracy (MOTA). It highlights that CTD consistently achieves superior accuracy levels at the same speed compared to the fixed frame skipping strategy (0\% confidence threshold), underscoring the efficacy of CTD. Furthermore, it offers valuable guidance for selecting an appropriate confidence threshold. By referring to Figure 3(c), one can determine the confidence threshold associated with the highest speed while maintaining the desired level of accuracy.

The analysis demonstrates superior accuracy levels compared to the fixed frame skipping strategy. Furthermore, it yields valuable insights into optimizing CTD's parameters for different tracking scenarios, enhancing its adaptability and effectiveness in real-world applications. 

\subsection{A Comparison between CTD with Different Trackers}

The CTD framework's adaptability was assessed across various detection models, such as YOLOv3, MobileNet, and SqueezeNet. According to the data in Table \ref{table:diff-detector}, a consistent speed-accuracy balance is maintained across all models when incorporated with CTD. Notably, a mere 10\% reduction in the confidence score threshold to 90\% achieves a significant speed enhancement of about 30\%, with only a negligible 3\% compromise in accuracy. When the confidence score threshold is further relaxed to 10\%, there is an impressive increase in speed by approximately 145\%, with an acceptable decrease in accuracy of 27\%. These outcomes highlight two principal advantages of the CTD framework: its plug-and-play compatibility with different platforms and models, and its ability to deliver substantial improvements in processing speed while sustaining satisfactory accuracy levels.

\section{Conclusion}
\label{sec:conclusion}

This work presents the CTD approach, a novel technique that significantly improves tracking speed while maintaining accuracy in tracking-by-detection systems. The CTD leverages the tracker's confidence score to strategically activate object detection. This paper offers three key contributions:

\begin{itemize}
    \item \textbf{Enhanced Speed:} CTD demonstrably increases tracking speed compared to conventional methods by selectively skipping detection in frames with low confidence scores.
    \item \textbf{Optimal Accuracy and Speed Trade-Off:} CTD achieves an optimal tradeoff between accuracy and speed through comprehensive experimentation. The experiments also provide valuable guidance for optimizing CTD in real-world deployments.
    \item \textbf{Adaptable Framework:} CTD has demonstrated great robustness and versatility through extensive experiments across a range of detection models.

\end{itemize}
By achieving a significant speedup with minimal accuracy loss, the adaptable framework CTD paves the way for real-time object tracking in resource-constrained environments.

\renewcommand{\bibfont}{\footnotesize}

\footnotesize{
\bibliographystyle{IEEEtran}
\bibliography{main}

\begin{thebibliography}{10}
\providecommand{\url}[1]{#1}
\csname url@samestyle\endcsname
\providecommand{\newblock}{\relax}
\providecommand{\bibinfo}[2]{#2}
\providecommand{\BIBentrySTDinterwordspacing}{\spaceskip=0pt\relax}
\providecommand{\BIBentryALTinterwordstretchfactor}{4}
\providecommand{\BIBentryALTinterwordspacing}{\spaceskip=\fontdimen2\font plus
\BIBentryALTinterwordstretchfactor\fontdimen3\font minus \fontdimen4\font\relax}
\providecommand{\BIBforeignlanguage}[2]{{%
\expandafter\ifx\csname l@#1\endcsname\relax
\typeout{** WARNING: IEEEtran.bst: No hyphenation pattern has been}%
\typeout{** loaded for the language `#1'. Using the pattern for}%
\typeout{** the default language instead.}%
\else
\language=\csname l@#1\endcsname
\fi
#2}}
\providecommand{\BIBdecl}{\relax}
\BIBdecl

\bibitem{Bewley2016}
A.~Bewley, Z.~Ge, L.~Ott, F.~Ramos, and B.~Upcroft, ``Simple online and realtime tracking,'' \emph{arxiv 1602.00763}, 2016.

\bibitem{Leal-Taixe2016}
L.~Leal-Taixe, C.~Canton-Ferrer, and K.~Schindler, ``{Learning by Tracking: Siamese CNN for Robust Target Association},'' in \emph{CVPRW}, 2016.

\bibitem{Wojke2018}
N.~Wojke, A.~Bewley, and D.~Paulus, ``Simple online and realtime tracking with a deep association metric,'' \emph{2018 International Conference on Image Processing}, vol. 2017-Septe, pp. 3645--3649, 2018.

\bibitem{Xiang2015}
Y.~Xiang, A.~Alahi, and S.~Savarese, ``{Learning to track: Online multi-object tracking by decision making},'' in \emph{Proceedings of the IEEE International Conference on Computer Vision}, 2015.

\bibitem{Huang2015}
C.~H. Huang \emph{et~al.}, ``Toward user-specific tracking by detection of human shapes in multi-cameras,'' in \emph{Proceedings of the IEEE Computer Society Conference on Computer Vision and Pattern Recognition}, 2015.

\bibitem{Zaman_2023}
A.~Zaman \emph{et~al.}, ``Artificial intelligence-aided grade crossing safety violation detection methodology and a case study in new jersey,'' \emph{Transportation Research Record: Journal of the Transportation Research Board}, vol. 2677, no.~10, p. 688–706, May 2023.

\bibitem{Bouchrika2016}
I.~Bouchrika \emph{et~al.}, ``Towards automated visual surveillance using gait for identity recognition and tracking across multiple non-intersecting cameras,'' \emph{Multimedia Tools and Applications}, 2016.

\bibitem{Wojke2017simple}
N.~Wojke \emph{et~al.}, ``Simple online and realtime tracking with a deep association metric,'' in \emph{2017 IEEE International Conference on Image Processing}.\hskip 1em plus 0.5em minus 0.4em\relax IEEE, 2017, pp. 3645--3649.

\bibitem{peng2024maxk}
H.~Peng \emph{et~al.}, ``Maxk-gnn: Extremely fast gpu kernel design for accelerating graph neural networks training,'' in \emph{ASPLOS}, 2024, pp. 683--698.

\bibitem{Breitenstein2011}
M.~D. Breitenstein \emph{et~al.}, ``Online multiperson tracking-by-detection from a single, uncalibrated camera,'' \emph{IEEE Transactions on Pattern Analysis and Machine Intelligence}, 2011.

\bibitem{Kuo2013}
S.~M. Kuo, B.~H. Lee, and W.~Tian, \emph{Real-Time Digital Signal Processing: Fundamentals, Implementations and Applications}.\hskip 1em plus 0.5em minus 0.4em\relax Wiley, 2013.

\bibitem{Redmon2018}
J.~Redmon and A.~Farhadi, ``Yolov3: An incremental improvement,'' \emph{arxiv 1804.02767}, 2018.

\bibitem{Girshick2014}
R.~Girshick, J.~Donahue, T.~Darrell, and J.~Malik, ``Rich feature hierarchies for accurate object detection and semantic segmentation,'' in \emph{Proceedings of the IEEE Computer Society Conference on Computer Vision and Pattern Recognition}, 2014.

\bibitem{Howard2017}
A.~G. Howard \emph{et~al.}, ``Mobilenets: Efficient convolutional neural networks for mobile vision applications,'' \emph{arxiv 1704.04861}, 2017.

\bibitem{Liu2016}
W.~Liu \emph{et~al.}, ``{SSD: Single shot multibox detector},'' \emph{Lecture Notes in Computer Science (including subseries Lecture Notes in Artificial Intelligence and Lecture Notes in Bioinformatics)}, vol. 9905 LNCS, pp. 21--37, 2016.

\bibitem{Simonyan2015}
K.~Simonyan and A.~Zisserman, ``Very deep convolutional networks for large-scale image recognition,'' \emph{International Conference on Learning Representations (ICRL)}, 2015.

\bibitem{Krizhevsky2012}
A.~Krizhevsky, I.~Sutskever, and G.~E. Hinton, ``Alexnet,'' \emph{Advances In Neural Information Processing Systems}, 2012.

\bibitem{Chang2014}
G.~Chang, ``Robust kalman filtering based on mahalanobis distance as outlier judging criterion,'' \emph{Journal of Geodesy}, 2014.

\bibitem{smalheiser2021effect}
N.~R. Smalheiser, E.~E. Graetz, Z.~Yu, and J.~Wang, ``Effect size, sample size and power of forced swim test assays in mice: Guidelines for investigators to optimize reproducibility,'' \emph{PloS one}, vol.~16, no.~2, p. e0243668, 2021.

\bibitem{Pearson1992}
K.~Pearson, \emph{On the Criterion that a Given System of Deviations from the Probable in the Case of a Correlated System of Variables is Such that it Can be Reasonably Supposed to have Arisen from Random Sampling}.\hskip 1em plus 0.5em minus 0.4em\relax Springer New York, 1992, pp. 11--28.

\bibitem{wang2023embracing}
H.~Wang, Z.~Tang, S.~Zhang, C.~Shen, and T.-H. Chang, ``Embracing uncertainty: A diffusion generative model of spectrum efficiency in 5g networks,'' in \emph{2023 International Conference on Wireless Communications and Signal Processing (WCSP)}.\hskip 1em plus 0.5em minus 0.4em\relax IEEE, 2023, pp. 880--885.

\bibitem{tang2024zerothorder}
Z.~Tang, D.~Rybin, and T.-H. Chang, ``Zeroth-order optimization meets human feedback: Provable learning via ranking oracles,'' in \emph{The Twelfth International Conference on Learning Representations}, 2024.

\bibitem{wang2021deminet}
Y.~Wang \emph{et~al.}, ``Deminet: dependency-aware multi-interest network with self-supervised graph learning for click-through rate prediction,'' \emph{arXiv preprint arXiv:2109.12512}, 2021.

\bibitem{MOTChallenge2015}
L.~Leal-Taix\'{e}, A.~Milan, I.~Reid, S.~Roth, and K.~Schindler, ``{MOTC}hallenge 2015: {T}owards a benchmark for multi-target tracking,'' \emph{arXiv:1504.01942}, 2015.

\bibitem{TSDA_OAL}
D.~Han, J.~Ju, H.~ko, D.~Kim, and B.~Ku, ``Online multi-person tracking with two-stage data association and online appearance model learning,'' \emph{IET Computer Vision}, vol.~11, 07 2016.

\bibitem{TC_SIAMESE}
Y.~Young~Chul \emph{et~al.}, ``Online multi-object tracking using selective deep appearance matching,'' in \emph{2018 IEEE International Conference on Consumer Electronics - Asia (ICCE-Asia)}, 06 2018, pp. 206--212.

\bibitem{GMPHD}
Y.-M. Song, K.~Yoon, Y.-C. Yoon, K.~C. Yow, and M.~Jeon, ``Online multi-object tracking with gmphd filter and occlusion group management,'' \emph{IEEE Access}, vol.~7, pp. 165\,103--165\,121, 2019.

\end{thebibliography}
}

\end{document}